# FlexParser—The adaptive log file parser for continuous results in a changing world

Nadine Rücker | Andreas Maier

Pattern Recognition Lab, Department of Computer Science, Friedrich-Alexander-Universität Erlangen-Nürnberg, Martensstr. 3, Erlangen, 91058, Germany

**Correspondence**
Nadine Rücker, Pattern Recognition Lab, FAU Erlangen-Nürnberg, Erlangen, Germany.
Email: nadine.ruecker@fau.de

**Abstract**

Any modern system writes events into files, called log files. Those contain crucial information which are subject to various analyses. Examples range from cybersecurity, intrusion detection over usage analyses to trouble shooting. Before data analysis is possible, desired information needs to be extracted first out of the semi-structured log messages. State-of-the-art event parsing often assumes static log events. However, any modern system is updated consistently and with updates also log file structures can change. We call those changes "mutation" and study parsing performance for different mutation cases. Latest research discovers mutations using anomaly detection post mortem, however, does not cover actual continuous parsing. Thus, we propose a novel and flexible parser, called *FlexParser*, which can extract desired values despite gradual changes in the log messages. It implies basic text preprocessing followed by a supervised Deep Learning method. We train a stateful LSTM on parsing one event per data set. Statefulness enforces the model to learn log message structures across several examples. Our model was tested on seven different, publicly available log file data sets and various kinds of mutations. Exhibiting an average F1-Score of 0.98, it outperforms other Deep Learning methods as well as state-of-the-art unsupervised parsers.

## 1 | INTRODUCTION

The number and size of running systems are growing every day and with it the number of log files stored. System log files contain various information which is written during system's operation. Their structural content is defined with software development and stored as text. The log messages can hold warnings, errors, and information about events. Common variables being logged range from key performance indicators (KPIs), over user behavior to transactions. Thus, log files are subject to great potential in the fields of anomaly detection,[1–3] failure monitoring,[4] usage analysis,[5] and many more. Manifold knowledge can be derived from logs in order to perform data-driven decisions. The very first step is to extract desired information and transform it into structured tables, also called parsing. In this work, we focus on event parsing where crucial information (the value belonging to an event) is extracted out of log files.

With the vast amount of information collected over time, manual exploitation becomes a tedious chore and is replaced by automated processing. Modern cloud computing systems, for example, produce up to 200 million log lines per hour.[6] As soon as the log structure is found, parsing rules can be set up and relevant information can be extracted automatically. However, as systems and their applications are continuously developed further, also log event structures are subject to change. Yuan et al[7] studied logging practices and logging modifications. They found log messages to speed up diagnosis time, however, exhibited that logging code is actively maintained by developers and undergoes constant changes. Logging code was found to be modified in over 18% of all committed revisions. In our studies, we call those modifications "mutations." For example, in Google systems high code churn rates cause that 36% of all log messages are subject to one or more changes throughout their lifetime.[8] Thus, automated log parsing that relies on static structures will result in incomplete data sets. Following, all subsequent data analyses are impacted and lead to wrong conclusions in the worst case. As an example, imagine a mobile health app tracks and logs the user's number of steps. The log message contains "Number of steps: 10.000." Automated parsers are set up to extract the number written after the key words "Number of steps." Continuous software development and improvement might trigger a logging statement correction. Subsequently, this log message might change with the next software update to "Number of strides: 10.000." The old parsing pattern does not apply any more and no value is parsed nor propagated to subsequent data analyses. This results in reduced data quality, wrong step tracking and user frustration. False notifications to the user indicating that the daily goal has not been reached can emphasize the negative product experience and decrease manufacturer reputation. In cyber security applications, even more critical consequences can follow. Constant software evolution and the resulting need for automated parsers that are flexible to logging changes are found in all domains.

In our research, we tackle the problem of correctly parsing ever-changing log messages in an automated way. We employ Deep Learning methods to build a flexible parser which adapts to gradual changes in the log file events. Following, we enable reliable parsing throughout software evolution and pave the way for successful log mining.

## 2 | RELATED WORK

Correct parsing lays the foundation for all subsequent data analyses. One pre-dominant log file parsing method is to turn expert's knowledge about which events to look for into regular expressions.[9,10] However, this is a highly manual process and previous research emphasized the importance and relevance of automated parsers.[8,11,12] The idea of using a hierarchical parsing system that first decides for a log file type and subsequently performs actual parsing has been protected by a running US patent called *Dynamic parsing rules*.[13] Furthermore, in order to determine crucial events automatically, frequent pattern mining[14] or artificial intelligence for file pattern analysis[11,15] were applied. Since log files pose a continuous stream of information, also Hidden Markov Models (HMMs) were employed for failure monitoring.[5] The research group LogPAI (Log Analytics Powered by AI)[11,15] has been providing a logparser toolkit which covers several different state-of-the-art log parsers. Those parsers extract message templates from log files automatically. One of the approaches is called Length Matters Clustering (LenMa).[16] It finds templates by utilizing the token length properties of log messages. Another clustering based approach was introduced under the name Iterative Partitioning Log Mining (IPLoM)[17] which partitions log data in a three-step hierarchical partitioning process into desired clusters. IPLoM outperforms other state-of-the-art methods as it does not depend on pattern frequency and thus, finds rare as well as highly frequent patterns, similarly. Furthermore, Log Key Extraction (LKE)[12] is a machine learning approach to cluster and detect anomalies in log files which is based on Finite State Automaton (FSA). Message templates can also be extracted by abstracting the log messages, which is exactly the underlying idea of an approach called Abstracting Execution Logs (AEL).[18] The method categorizes similar log messages and merges events which have only a set number of different tokens. Thus, this method's performance depends on an appropriate, data-set specific threshold. Another Log File Abstraction (LFA)[19] method utilizes token frequencies within each log message and extracts log events for all log messages, independently. Token frequency is also the basis for LogSig[20] which categorizes log messages into a set of event types. Another state-of-the-art method that extracts high quality patterns from log messages is called LogMine.[21] It exhibits high computational efficiency, can process millions of log files within seconds and extracts patterns reliably. It leverages the fact that log files contain stable structures due to their automatic generation. Moreover, Streaming Parser for Event Logs (Spell)[22] mimics human instinct of finding the longest common subsequence in order to find event patterns in log messages. Next to clustering and abstraction methods, also online, tree-based approaches show effective results. For example, in SHISO[6] a structured tree is created using the nodes generated from log messages and refines the result incrementally in real-time. He et al. proposed a similar method called Drain[23] which accelerates the parsing process by using only balanced and fixed depth parse trees. Furthermore, evolutionary approaches were employed, such as Multi-objective Log message Format Identification (MoLFI).[24] It is a search-based approach based on the multi-objective genetic algorithm, NSGA-II.[25]

In this work, we applied Deep Learning methods to flexibly parse changing logs. Deep Learning has been studied widely and applied in many research fields such as medical imaging, speech recognition, object detection, genomics and many more.[26] Deep Learning abstracts the relation between input and output as several layers of processing. The combination of such layers in an artificial, neural network allows to learn complex relations.[27] Different types of neural networks evolved over the last years. Deep convolutional networks exhibit advantages in image, video, speech and other processing. Furthermore, sequential data such as text and speech require networks to learn temporal dynamic behavior. This has been addressed by recurrent neural networks (RNNs) which allow previous outputs to be used as inputs. However, a common problem in traditional RNNs has been the vanishing gradient. Two special kinds of RNNs called Long Short-Term Memory (LSTM)[28] and Gated Recurrent Unit (GRU)[29] tackle this problem. LSTMs were studied already in 1997 by Hochreiter et al[28] but have been employed widely yet more than 10 years later when access to higher computational power became easier. A common architecture of LSTM units contains one cell which memorizes information next to three gates, an input gate, an output gate, and a forget gate. Moreover, GRUs implement a special kind of LSTM which do not contain an output gate. Furthermore, Fully Convolutional Neural Networks (FCN) are end-to-end trained convolutional networks.[30] They first have been applied to images for semantic segmentation but also proved to solve time series classification problems successfully.[31] In order to improve the performance of an FCN, the model was augmented by either LSTM[32] or GRU sub-modules.[33]

In literature, log file processing has been studied from different angles and various methods have been applied. The need for automatic parsers which are adaptive to gradual changes has been described,[7] however, not yet addressed, holistically. For example, in Kuhnert et al,[34] the problem of incomplete parsing results was treated as a post-processing task. A classification method was employed in order to complete missing information in log files from old versioned software. Furthermore, with software development not only new information can be added to log files but also events can get modified or even dismissed. Thus, an adaptive solution using HMMs was proposed with the goal to parse texts automatically and flexibly.[35] However, their method was computationally expensive and has only been tested on one data set. Zhu et al[36] proposed a log-based anomaly detection method called LogRobust. They utilized semantic information out of log events, fed it to a Bi-LSTM model, and trained on the log events' importance. However, they validated the method on only two different data sets and did not cover changes in semantics. A similar approach is called Deeplog[3] where an LSTM model is trained on system logs treated as natural language sequences. It concentrated on detecting anomalies when log patterns deviate from the norm. However, Deeplog did not adapt patterns according to the detected changes in event log structures. Furthermore, SwissLog[37] is another Deep Learning based anomaly detection model which is robust to changing log data. These LSTM-based methods encountered major performance issues when processing unseen log events and values. Chen et al[38] addressed this problem by training LSTM models and introducing a novel transfer learning method. Unfortunately, they transferred only the anomalous knowledge, not the actual parsing of values. Concluding, all those methods only focused on detecting changes in log message structures. However, they did not cover the required adaption of parsing logic and assumed that the subsequent processing was manually updated to cover the changes. Our research fills this gap and addresses parsing values from gradually changing log messages in an application independent way using Deep Learning approaches.

## 3 | MATERIALS AND METHODS

For any content analysis based on any system's log files, crucial information needs to be extracted first. As those systems and with them the log file structures can change over time, we built an adaptive, flexible parser called *FlexParser*. First, we preprocessed the raw log files and reused existing, parsed values as labels which resulted from standard regular expression techniques. In order to demonstrate parsing challenges and the benefits of FlexParser, we concentrated on one crucial event type of interest per data set. Following, we leveraged supervised machine learning to train our models on one selected event per data set. This enabled the model to learn the standard structure and pattern for parsing the desired values but still allowed the algorithm to excuse mutations inside crucial log messages. In the following, we describe which data we trained and tested our algorithms on, how we preprocessed those before we present our Deep Learning models. Moreover, we explain our methodology to compare our results with state-of-the-art parsing performances.

### 3.1 | Data

We investigated our approach on different, publicly available log file data sets from loghub.[39] In order to demonstrate the general applicability of our approach, we chose seven different log file types which exhibit various structure, distribution, and frequency of desired events. The files were produced by distributed systems, supercomputers, operating systems as well as mobile systems. Table 1 holds data set details. Most log files follow some sort of structures. Typically all lines start with a date/time stamp and end with the actual event text. Some log files carry more information in between such as message types or IDs. Figure 1 shows the first lines of *HealthApp* log file as an example carrying three sections after the date/time stamp in each line. In our work, we focused on events in the event text which consist of a key (e.g., totalCalories) and a value (e.g., 126,775).

We trained our model on a sample set containing 6000 lines of log file entries ($X_{train}$) and on one single event per data set each. We created a ground truth ($y$) for training and validation by applying data set specific regular expressions. Afterwards, the extracted values were manually validated for all data sets and chosen events. Our model was set up to be trained separately for any desired event. This enabled parsing performance comparison across data sets and reduced result complexity while maintaining great data set variety. Nevertheless, scalability to parse several event-value pairs at once is given.

For testing, we used 2000 unseen log file messages. Consequently, distinct data sets were used for training (75%) and testing (25%). Table 2 holds per data set an example log file line which contains the event key and the desired event value. For each data set we chose one information of interest which shall be parsed. We manually selected the respective event such that use case diversity and general applicability can be

**TABLE 1** Log data set overview

| Data Set | System | Description | #Messages | Data Size |
|---|---|---|---|---|
| Spark | Distributed System | Spark job log | 33,236,604 | 2.75GB |
| BGL | Supercomputer | Blue Gene/L supercomputer log | 4,747,963 | 708.76MB |
| Windows | Operating System | Windows event log | 114,608,388 | 26.09GB |
| Linux | Operating System | Linux system log | 25,567 | 2.25MB |
| Mac | Operating System | Mac OS log | 117,283 | 16.09MB |
| Android | Mobile System | Android framework log | 1,555,005 | 183.37MB |
| HealthApp | Mobile System | Health app log | 253,395 | 22.44MB |

```
20171223-22:15:29:606|Step_LSC|30002312|onStandStepChanged 3579
20171223-22:15:29:615|Step_LSC|30002312|onExtend:1514038530000 14 0 4
20171223-22:15:29:633|Step_StandReportReceiver|30002312|onReceive action: android.intent.action.SCREEN_ON
20171223-22:15:29:635|Step_LSC|30002312|processHandleBroadcastAction action:android.intent.action.SCREEN_ON
20171223-22:15:29:635|Step_StandStepCounter|30002312|flush sensor data
20171223-22:15:29:635|Step_SPUtils|30002312| getTodayTotalDetailSteps = 1514038440000##6993##548365##8661##12266##27164404
20171223-22:15:29:636|Step_SPUtils|30002312|setTodayTotalDetailSteps=1514038440000##7007##548365##8661##12361##27173954
20171223-22:15:29:636|Step_LSC|30002312|onStandStepChanged 3579
20171223-22:15:29:645|Step_ExtSDM|30002312|calculateCaloriesWithCache totalCalories=126775
20171223-22:15:29:648|Step_ExtSDM|30002312|calculateAltitudeWithCache totalAltitude=240
20171223-22:15:29:649|Step_StandReportReceiver|30002312|REPORT : 7007 5002 150089 240
```

**FIGURE 1** Example log file lines from data set *HealthApp*

**TABLE 2** For each data set one log file line example is presented

| Log data set | Event frequency | Example line X | Event key | Event value y | Event key syn | Event key err |
|---|---|---|---|---|---|---|
| Spark | 15.3% | executor.CoarseGrainedExecutorBackend: Got assigned task 0 | task (311) | 0 (4) | duty (145) | tusk (145) |
| BGL | 1.5% | ciod: generated 64 core files for program /bgl/apps/swl-prep/ibm-swl/functional/sppm_chkpt/run/sppm | generated (241) | 64 (6) | created (29) | gennerated (29) |
| Windows | 27.9% | Read out cached package applicability for package: Package_for_KB3121255 31bf3856ad364e35 amd64 6.1.1.0, ApplicableState: 112, CurrentState:112 | ApplicableState (20) | 112 (3) | ApplState (6) | ApplicabelState (6) |
| Linux | 6.2% | combo su(pam_unix)[22644]: session opened for user news by (uid=0) | user (12) | news (4) | name (5614) | use (1020) |
| Mac | 0.8% | calvisitor-10-105-160-95 kernel[0]: AppleThunderboltGenericHAL::earlyWake - complete - took 0 milliseconds | took (91) | 0 (11) | take (91) | lasted (10) |
| Android | 1.3% | acquire lock=166121161, flags=0x1, tag="RILJ_ACK_WL", name=com.Android.phone, ws=null, uid=1001, pid=2626 | lock (122) | 166121161 (15) | fix (16) | lokk (16) |
| HealthApp | 12.1% | calculateCaloriesWithCache totalCalories=126775 | totalCalories (374) | 126775 (887) | totalCal (201) | totalCallory (201) |

*Notes*: Furthermore, the chosen event key and value pairs are provided. The entire log content per data set is stored in *X*. The event value found within the log message poses the to be parsed value and is stored per line in vector *y*. Furthermore, this table holds two mutations per key. One mutation follows the synonym of event called *Event Key Syn*, the other mutation exhibits spelling errors called *Event Key Err*. The numbers in brackets hold the respective token representation.

demonstrated. Variety is achieved in terms of several properties. These are, among others, event frequency within the data set, log line length, special character usage in between event key and value (e.g., "=", ":"), event value type, distribution, and variety. Table 2 holds event frequency within the first 2000 lines per data set in the second column and emphasizes data set variety. All those properties can affect parsing performance which our methods were tested on. The entire log file text was considered as *X*, whereas the to be parsed event values posed the desired result *y*. For an intuition about log file contents and differences, Figure 2 presents the plotted to be parsed values, *y*, found within the first 2000 lines per log data set. The default *y* per line is "False" indicating that no value shall be parsed from that line. Figure 2A illustrates that *y* of *Spark* exhibits highest frequency of event values and continuous gradient. On the contrary, *BGL*, *Mac* and *Android* contain sparse occurrence of event values. Figure 2C shows *y* of *Windows* which arise in bulk, whereas *HealthApp's y* value rises in the beginning and becomes stable toward end of the plot. Only a few distinct values of *y*, however at a medium to high frequency, are found in Figure 2D of *Linux*. In summary, these plots prove great variety across the studied data sets in terms of values, value types, distribution, and frequency. Thus, chosen data sets and to be parsed events serve the purpose of general parsing performance evaluation.

Furthermore, Table 2 holds two mutations of the event key. We introduced those to mimic changes in the log file content during system's lifetime. These mutations were chosen per data set in order to study log modifications according to their underlying root cause and mimic real-world coding behavior. Ninety-eight percent of all logging code modifications were found within the log message contents according to a study from Yuan et al.[7] Furthermore, they found that log modifications to text in log messages mostly originate from inconsistencies (39%), clarifications (36%) and spelling errors (18%). Software developers address the first two by replacing event keys with a better suiting string used in the logging code. Their intention for this modification is to add clarity while not changing the meaning. In order to represent those logging modifications in our studies, we chose a mutation called *Event Key Syn*. It represented the logging improvement by using a synonymous word as well as served to study possible semantic influences. This enabled parsing evaluation for cases where the same variables were logged under a different name, thus, event keys have been modified but not their meaning. Furthermore, in order to mimic both, modifications which adapt the word's meaning and spelling errors, we introduced another mutation kind called *Event Key Err*. That represented any event changes which can deteriorate parsing

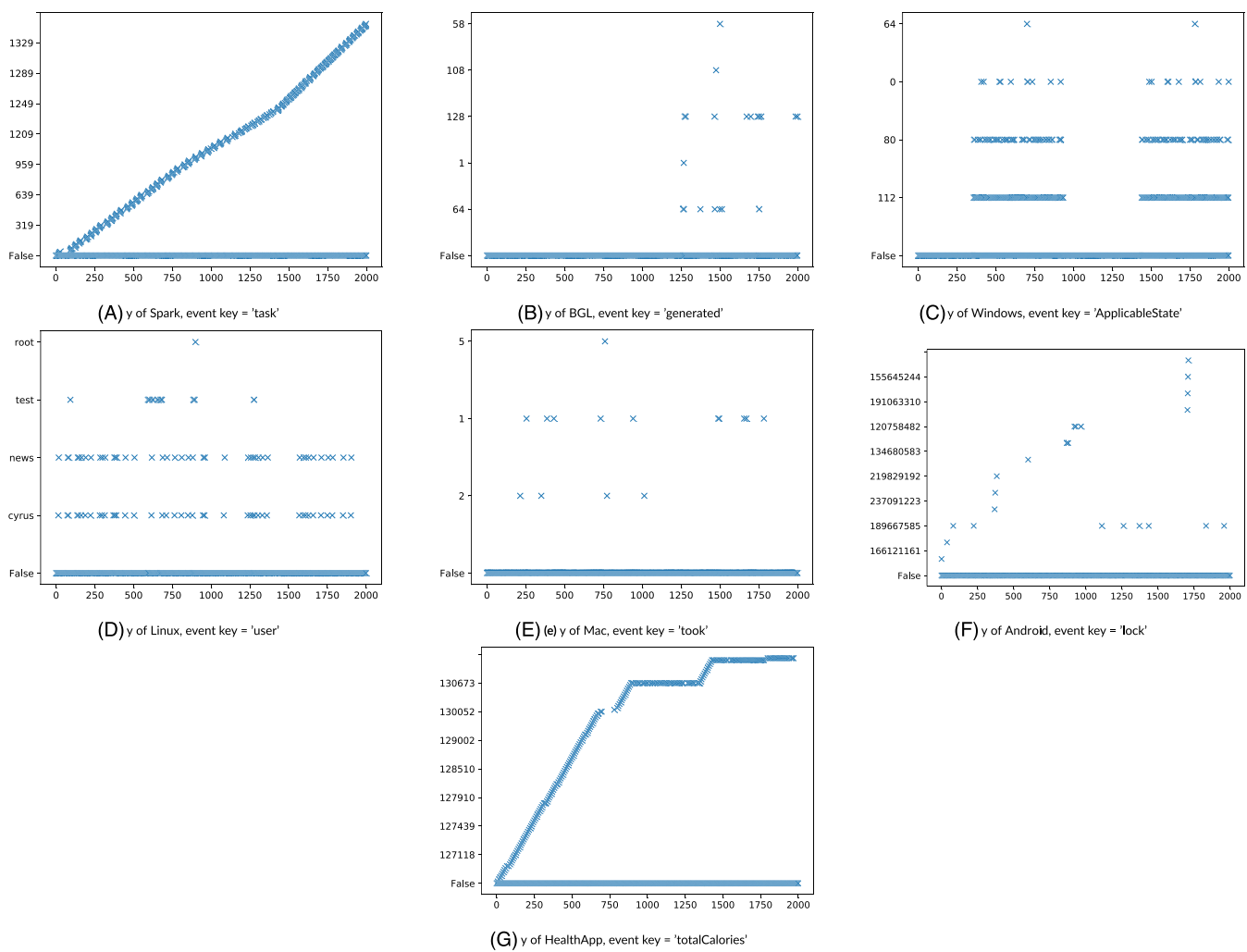

**FIGURE 2** Plots of *y* (the to be parsed event value) per data set and per log line. *y* was set to "False" in case the respective line did not contain the desired information. The first 2000 lines of each log data set are displayed

template matching and NLP based recognition. We generated these mutations by replacing the event key with its respective *Event Key Err* or *Event Key Syn*. Both kinds of parsing approaches studied in this work learn how to parse depending on event frequencies. In order to study various time points of mutation happening and, thus, different distributions of mutant and non-mutant proportions, we induced mutation starting from three different line numbers 500, 1000, and 1500. Hence, we tested our flexible parsing on data sets where 75%, 50%, and 25% of all considered log messages have undergone mutation. Thus, we worked with seven versions per data set including one non-mutated version, and six more sets containing *Event Key Err* and *Event Key Syn* mutation each applied starting at three different line numbers.

## 3.2 | Preprocessing

In order to establish a model which parses desired information from log files, data needed to be preprocessed first. All steps were applied to the data sets individually. Our data sets contained semi-structured text which was stored line by line in vector *X*. Furthermore, vector *y* held the desired parsing result per line. Per default, the elements of *y* were set to "False" if the respective line did not contain an event value. Thus, *X* and *y* exhibited same length. We started with normalizing the corpus, followed by tokenization. In the third step tokens were vectorized. All steps are described in detail in the following and were applied to *X* and *y* equally. This ensured that the same values found in raw *X* and *y* kept their relation after preprocessing and could be parsed correctly.

Figure 3 visualizes preprocessing of an exemplary log message from the *HealthApp* dataset. The first task of corpus normalization was to remove words which occur very frequently, however, did not add value in information retrieval tasks. These words are called stop words. As all our log file data was written in English, we leveraged Python's stop word list for English text from its natural language processing toolkit. Before we applied stop word removal, we excluded "no" and "not" from the list to preserve negations. In our exemplary log message no stop words were found, thus, no words were removed. Following, we excluded special characters and turned all letters to lower case. Digits were preserved as most of our data sets contained numeric event values which posed the desired parsing results. Moreover, we applied lemmatization which takes part of speech into account and reduces single, even inflected words to the common, main group of meaning.

Next, we converted the extracted elements into tokens which are displayed in Figure 3 in the third box from the top. As the length of log file lines varied, we padded and truncated each line for further processing. The desired line length was determined by the maximum number of elements of those lines which contained an event value other than "False." This ensured that no crucial event keys and values were truncated while cropping not needed excess text. Following, each element was uniquely assigned to an integer. We did not use existing encoding models since log file contents are very domain specific, do not mirror natural language and, thus, are not well represented by existing models. Following, we pre-trained one encoding model per data set. Therein, we followed the state-of-the-art tokenization approach. The higher an element's frequency, the lower integers are assigned to it, respectively. For our use case, we set a maximum number of 10,000 words for our vocabulary. All elements which occurred less often than the most frequent 10,000 elements were assigned to the same out of vocabulary token. In order to prevent that desired event values were outside of the vocabulary and impossible to distinguish by the model, we assigned high enough frequencies to those to be parsed values found in y.

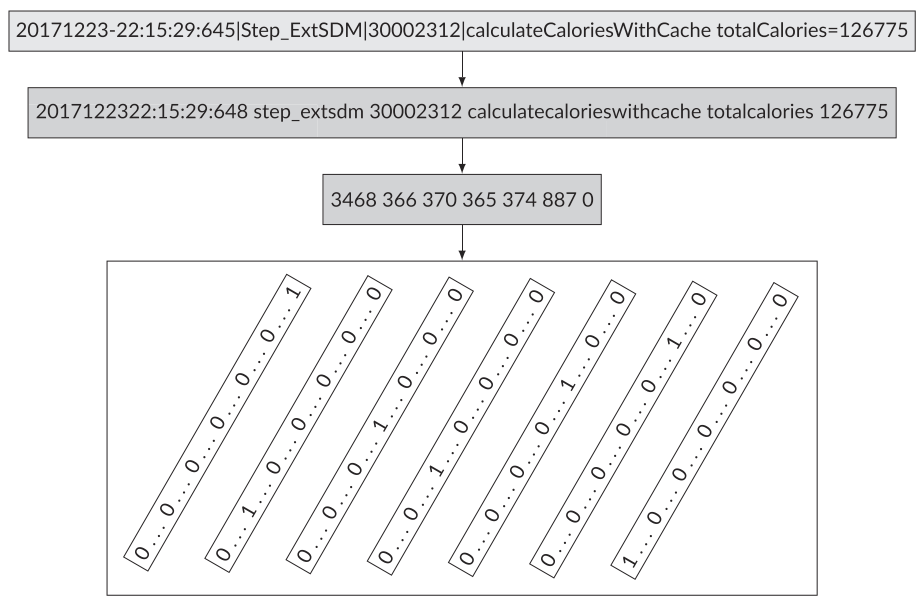

**FIGURE 3** Tokenization and vectorization of an event message. In the upper box, one can see an original log file line ($X_i$) from *HealthApp*. The second box contains the event messages after normalization. In the following step strings were split into tokens. The third box visualizes the individual event log tokens translated into their respective integer representation. Thus, the key value "126775" found in the original log file line, for example, was represented by the token "887." The event key "*totalCalories*" was turned into the token "374." Finally, tokens were vectorized using one-hot-encoding which is indicated in the lowest box

The final preprocessing task contained transformation of tokens into high-dimensional vectors as we worked with categorical data. The desired parsing result must be exactly equal to the event value given in the raw log file; thus, we could only receive a true or false parsing result. Therefore, we applied one-hot-encoding which represented tokens in a binary representation and spanned a result space where, for example, token 401 is as distant from 400 as 2000. This step implicitly assured that the exact integer assigned per token did not make any difference. After preprocessing the log file text (X) and desired parsing values (y), Deep Learning models could be trained on those vectors to learn which value poses the correct parsing result from a given log message.

## 3.3 | Parsers

We propose *FlexParser*, a novel, Deep Learning based approach to parse log file texts effectively despite gradual changes in crucial lines of text. It comprises text preprocessing according to techniques presented in Section 3.2 and a stateful LSTM model. In the following, we present the model architecture of the stateful LSTM as well as other parsing options which we employ for comparison. We applied five different Deep Learning parsers (DL-parsers). After preprocessing 8000 log lines, we trained our DL models on a subset of 6000 lines per data set. Next, we tested our trained model on the remaining, preprocessed 2000 lines of logs and, thus, worked on separate training and test sets. The Deep Learning models were trained on the non-mutated log messages only and are described in the following. All hyperparameters were determined using grid search hyperparameter tuning while optimizing the F1-Score.

In addition, we trained and tested eleven traditional, pattern-based log parsers (see Section 2). The individual found parsing patterns were updated for all mutation cases and proportions. Thus, we implemented and evaluated five DL-parsers and eleven state-of-the-art log parsers (SOTA-parsers). *FlexParser* as one of the studied Deep Learning methods poses the proposed approach.

### 3.3.1 | Long short-term memory

The first model we trained was a basic LSTM with 32 units. We followed the intuition that a parsing pattern can be determined by very recent structures but also by decisive information which occurred longer time ago. This combination was used to prevent parsing failure in case of mutated log files as the entire sequence is taken into account. In order to prevent overfitting, we applied local average pooling followed by a drop-out operation with rate 0.61. The final result was calculated by a dense layer with softmax activation function. For our work, we treated one log message as a sequence where each token posed one time step. Per token (i.e., time step) was the feature vector spanned which contained the one-hot-encoded version of the token. We trained the model optimizing by F1-Score until convergence.

### 3.3.2 | Stateful Long Short-Term Memory

The standard implementation of LSTM implies that states are reset after each training epoch. Thus, the trained weights will depend on all seen sequences and contain information about all log messages, implicitly. However, the direct relation across sequences is not kept. This can result in a model learning to map sequences to their exact label, while not learning the actual parsing pattern. Thus, we enhanced the basic LSTM's learning structure and enabled statefulness. Consequently, the parallel trained states were not reset after each epoch but were reused and refined over the course of several epochs. In our experiments, we found 40 epochs to benefit the learning process in all data sets. On average over all data sets, 40 epochs contained five events which held crucial information to be parsed. This resulted in the model learning the actual parsing logic as depicted in Equation (2) in contrast to learning one-to-one relation (see Equation (1)).

$$\begin{aligned}&\text{"the interesting value}=123,\ \text{some other text,"\ label}:\text{"123"}\rightarrow\text{learn to map this entire stirng to 123}\\&\text{"another value}=456,\ \text{some more text,"\ label}:\text{"456"}\rightarrow\text{learn to map this entire string to 456}\end{aligned} \quad (1)$$

$$\left.\begin{aligned}&\text{"the interesting value}=123,\ \text{some other text,"\ label}:\text{"123"}\\&\text{"another value}=456,\ \text{some more text,"\ label}:\text{"456"}\end{aligned}\right\}\rightarrow\text{learn to extract the number between "value" and "some"} \quad (2)$$

### 3.3.3 | Fully convolutional network

Another state-of-the-art time series classification method is represented by the fully convolutional neural network (FCN).[30] Following the suggestion of Wang et al[40] we used three convolutional blocks. As depicted in Figure 4A, the first block held one convolution with 16 filters, batch normalization (BN) and sigmoid activation. The following convolutional blocks contained the same elements, but used 16 and 32 filters. The kernel sizes of the convolutional blocks were determined as 7, 5, and 5, respectively. Global average pooling and a dense layer using softmax activation finalized the setup. We used the Adam optimizer with learning rate $10^{-1}$ according to the hyperparameter tuning results.

### 3.3.4 | Long Short-Term Memory fully convolutional network

Karim et al[32] leveraged the benefits of LSTM and FCN and combined both in a model called LSTMFCN. Also, in our earlier research,[41,42] LSTMFCN posing a FCN with LSTM sub-modules outperformed both individual methods, respectively. Figure 4B illustrates the model setup. The left part holds the FCN with three subsequent, convolutional blocks of filter size 32, 32, and 512 followed by global pooling. Kernel sizes of 7, 5, and 5 were found to achieve the best F1-Scores. In parallel, a basic LSTM setup with 150 units was applied using a dropout rate of 0.17 to prevent overfitting. This LSTM block was followed by another dropout operation of rate 0.25. Both strands were concatenated before softmax represents the last component. We trained the model with the Adam optimizer using a learning rate of $10^{-2}$.

### 3.3.5 | Gated recurrent unit fully convolutional network

Furthermore, FCN can be augmented by models other than LSTM. As GRU showed advantages over LSTM in some applications, we also implemented FCN combined with a GRU module. The respective illustration is presented in Figure 4C. Hyperparameter tuning resulted in the same configuration as found to be best with respect to achieved F1-Score for LSTMFCN. The optimal value for GRU units was determined to be 32. In order to prevent overfitting, the GRU module included a dropout rate of 0.02 and recurring dropout rate of 0.25. The following dropout layer with rate 0.82 was determined to result in best performances according to hyperparameter tuning.

### 3.3.6 | Application of traditional log parsers

For comparison, we also applied traditional, state-of-the-art log parsers (SOTA-parsers) which extracted desired values out of log messages. We chose eleven different methods called AEL,[18] Drain,[23] IPLoM,[17] LenMa,[16] LFA,[19] LKE,[12] LogMine,[21] LogSig,[20] MoLFI,[24] SHISO,[6] and Spell[22] following the logpai[15] implementation. For an introduction, see Section 2. These are all unsupervised machine learning methods which learn a log file pattern based on log content frequencies and result in event templates. Figure 5 illustrates the common log parsing pipeline. It visualizes the log message creation step and results in the found event template per line. The event template contains information about variable and static parts. We postprocessed those results, filtered to the selected event according to Table 2 and extracted the desired event value by applying the event template as a regular expression. This yielded a vector $y_{pred}$ comparable to the same ground truth which was used in testing the Deep Learning models.

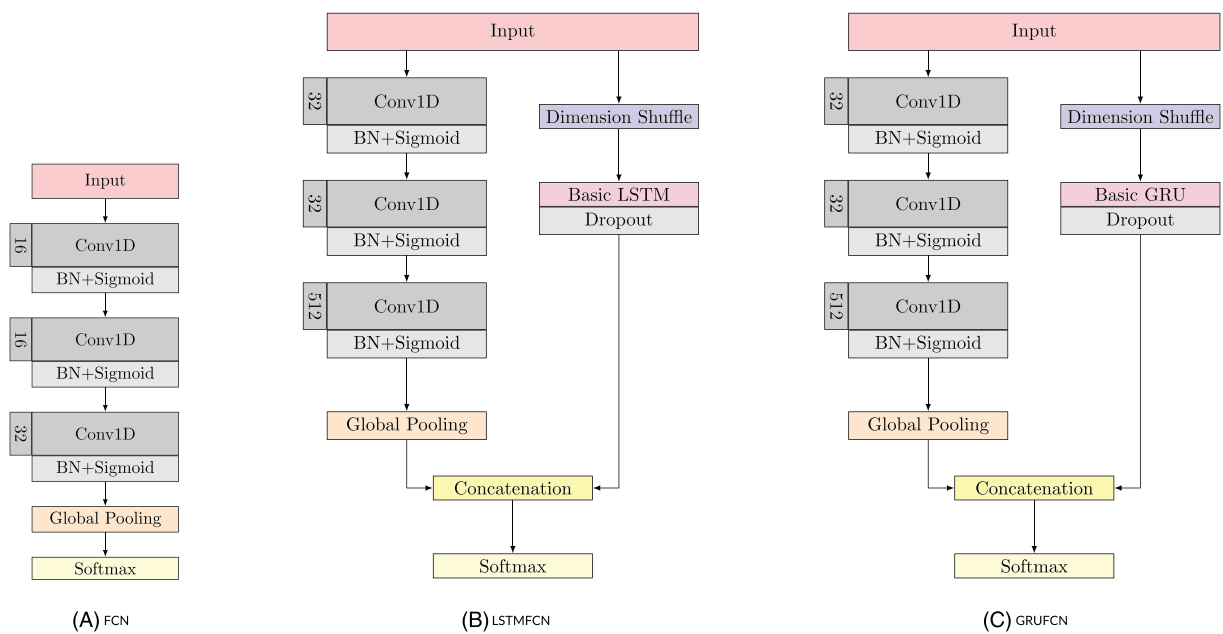

**FIGURE 4** Architecture of three different Deep Learning models

The advantage of those unsupervised methods over supervised Deep Learning approaches is shown in the ease of retraining as only the raw log files and no labels are required. Thus, in case of mutating and gradually changing log message event names, the methods could simply be trained again to adapt the event templates. Thus, for fair comparisons of resulting parsing performances, we trained the traditional log parsers on each mutation version separately. This resulted in adapted event templates for the mutation cases and updated regular expressions which we used for testing and evaluating parsing performance.

## 4 | RESULTS

We propose a log parsing method which learns log message structures in a flexible way such that correct values are still parsed despite log message mutations. Thus, we tested five Deep Learning methods and compared their results with those achieved by eleven SOTA log parsers. The decisive metric is the F1-Score as we compared a ground truth vector containing desired values with the parsing results of our models. Since most data sets exhibited imbalance, F1-Score holds more information than accuracy. In order to study parsing performance in case of mutating inputs, we introduced two different types of mutations which start at three various lines in the test sets, respectively. Furthermore, to achieve as general results as possible, we tested all models on seven distinct data sets which are described in Section 3.1. Per data set, we selected one event key of interest in order to reflect diversity and enable comparability.

### 4.1 | Non-mutated case

First, we present results of parsing log files which exhibit static template structures, referred to as the non-mutated case. Thus, we discuss the parsing capabilities of different models on several data sets where the respective training and test sets exhibit the same event keys and log structures. Figure 6 shows the parsing performance of all methods applied to all non-mutated test sets next to each other. The different models are depicted on the x-axis. Per model, the seven F1-Scores reached for seven data sets are illustrated using a box plot. We indicate the median F1-Score over all data sets per model by a light, horizontal bar. The models drawn from left to right follow the ascending order of their respective median F1-Score. Circles illustrate outliers.

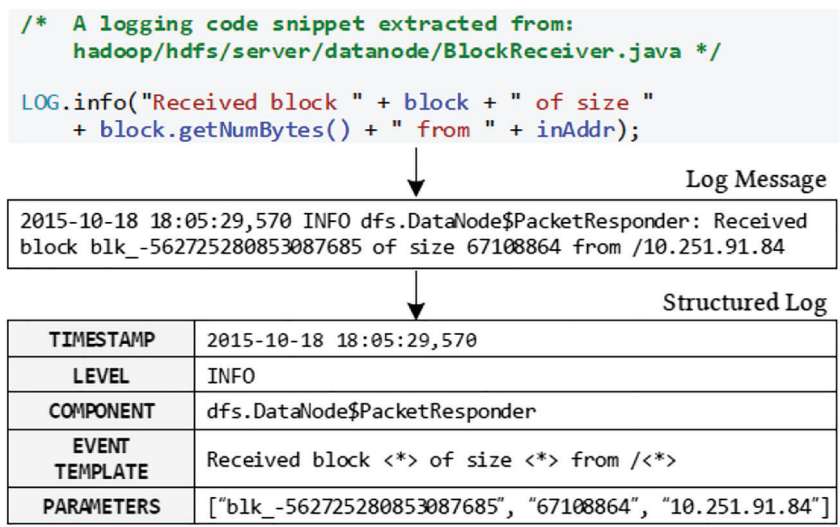

**FIGURE 5** The common log parsing pipeline which was shared by all here applied traditional log parsers.[15] The first part shows an exemplary code snippet which creates a log message line. It writes static and variable parts which are depicted in the second block. Parsers process all log messages and create a structured log which contains event templates per line

In the non-mutated test case, the trained models were applied to test sets which contain the same log patterns as the training set exhibits. This is also the reason for the fact that two SOTA-parsers, AEL and Drain, both result in the highest possible median F1-Score of 1.0. Both found the correct pattern for all datasets except one. The respective derived regular expressions fit to all log lines of interest. The outlier showing an F1-Score of 0.14 is depicted by a circle and was achieved for data set *Android*. This can be explained by the special properties of *Android*. The chosen event key was found in a log line featuring seven event values which all are varying parts (see Table 1). Furthermore, the log messages of interest were very sparse representing only 1.3% of all lines as illustrated in Figure 2F. Thus, when applying the SOTA-parsers with their standard configurations, the desired event value was only recognized as the value to be parsed when occurring with specific other event values in the same line in combination. Most of the SOTA-parsers showed parsing results highly depending on the applied data set (see Figure 6). Both, LogSig and Lenma, were not convincing for any data set. Other studies showed that they could find log file structures correctly.[16,20] However, in our setup, the found patterns could not be turned into successfully working regular expressions for actual parsing using the standard configurations. Manual adaptions of thresholds would have been necessary which penalized our goal of an automated setup.

From right to left, the best results represented by Drain and AEL are followed by all our studied Deep Learning models (stateful LSTM, FCN, GRUFCN, LSTMFCN, LSTM). They were able to learn correct parsing rules even for *Android* resulting in F1-Scores between 0.72 and 1.0 (see Figure 6). Among the DL-parsers, the lowest F1-Score was achieved when parsing *Windows* log files with LSTMFCN. In that case, the model overfitted on two predominant event values (80 and 112) which were found in over 95% of lines containing desired event values (see Figure 2C). We strive for a flexible parser, which is capable to parse all kinds of log file types. We found stateful LSTM to perform best with respect to mean (0.99) and median (1.00) F1-Score over all explored data sets. Statefulness is the only difference to the other tested LSTM. By leveraging the power of statefulness, we assured that the actual structure and pattern of entire log file lines were learnt and not only the relationship of specific log message to parsing result. In the example of *HealthApp*, the model learned that the value after "calculateCaloriesWithCache totalCalories=" was the desired one. On the contrary, it did not learn that "126797" belonged to "calculateCaloriesWithCache totalCalories=126797," as it was the case for the normal LSTM. Thus, the stateful LSTM is more flexible and adaptive to unseen values.

Overall, the F1-Scores of all DL-parsers held values above 72%, whereas the performance of remaining state-of-the-art methods highly depended on the data set and exhibited larger variance. Concluding, the studied DL-models can compete with SOTA-parsers already in the non-mutated case.

## 4.2 | Mutated case

Modern systems and their produced log files are subject to continuous evolution.[8] Following, we studied the parsing capabilities of our models when applied to mutated data sets as described in Section 3.1. Therein, we induced two kinds of mutations, each starting at three different log file

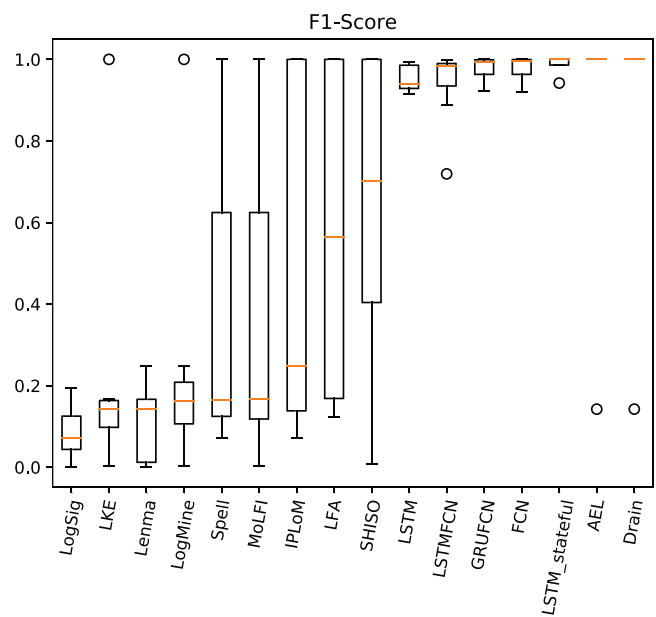

**FIGURE 6** Parsing performance per model for non-mutated case. Box plots contain F1-Scores for all seven tested data sets. The light horizontal bar indicates the median F1-Score over all test runs. Models on the x-axis are sorted by median F1-Score presenting the best model on the far right

lines. This resulted in various proportions of non-mutated and mutated log file lines to be parsed. Results for the mutation cases "syn" and "err" are illustrated in Figures 7 and 8, respectively. F1-Scores per model and mutation induction line are illustrated using box plots. Each box holds the seven F1-Scores for the seven studied data sets. Results for all three mutation induction lines are depicted per model and exhibit significant deterioration for all state-of-the-art parsers compared to the non-mutation case. However, all five deep-learning models (LSTM, LSTMFCN, GRUFCN, FCN, and stateful LSTM) exhibit no or only slight decrease in performance. Independent of the starting line of mutation, all DL-methods outperform SOTA-parsers with respect to median F1-Score. Among SOTA-parsers only, Drain and AEL remain ahead.

Detailed discussion of the results is organized into three parts. First, the two mutation cases ("syn" and "err") are evaluated and their influence on parsing capabilities is examined. Moreover, model performance on mutating log files is investigated. The third part is dedicated to compare the best state-of-the-art algorithm with our proposed DL-method, FlexParser.

### 4.2.1 | Mutation case comparison

We studied two different mutation types. One changed the event key into its synonym ("syn"), the other mutation imitated a spelling error ("err"). These represented text modifications excluding insertion and deletion. Comparing results for both types enables to examine influence of text preprocessing. For DL-parsers, we first applied several steps to prepare the text for training the models. It also contained lemmatization, which could already catch minor mutations. For example, for *Mac*, simulating a synonym change of "took" into "take" resulted in the same token representation "91" (see Table 2). However, in most cases the mutations were out of vocabulary (e.g., see *Android*, *HealthApp* in Table 2) or were tokenized because the mutation word coincidentally matches other log contents (see *Linux* in Table 2). Following, parsing results

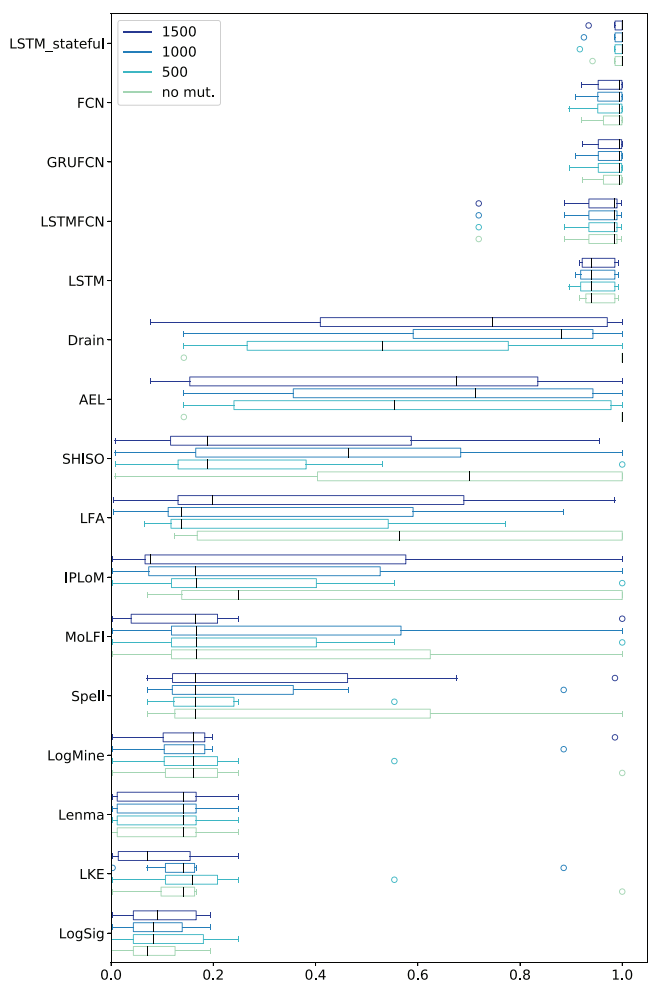

**FIGURE 7** Parsing performance achieved for the mutated case "syn." All models are drawn onto the vertical axis and hold four box plots each. The lightest box contains results of the non-mutated case for convenient comparison. The other box plots hold performances for the start of mutation in different line numbers. The box plots contain F1-Scores of all seven evaluated data sets. Outliers are indicated with circles, the black vertical bars plot the median F1-Score

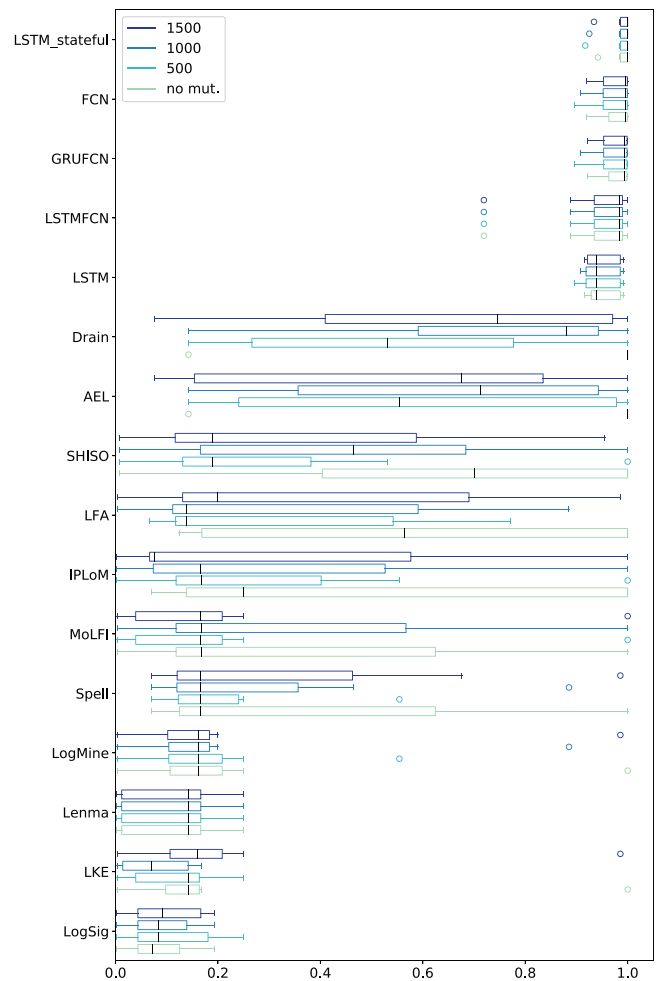

**FIGURE 8** Parsing performance achieved for the mutated case "err." All models are drawn onto the vertical axis and hold four box plots each. The lightest box contains results of the non-mutated case for convenient comparison. The other box plots hold performances for the start of mutation in different line numbers. The box plots contain F1-Scores of all seven evaluated data sets. Outliers are indicated with circles, the black vertical bars plot the median F1-Score

(see Figures 7 and 8) show that the trained DL-methods generalize already very well as they abstract the entire event key-value surrounding and do not rely on individual token triggers. Thus, influence of processing techniques which target semantics are negligible. As mutations often destroy semantic recognition, semantic independence poses a FlexParser benefit. Furthermore, preprocessing was still crucial to turn text into trainable vectors. All tokens were one-hot-encoded, thus, exhibited equal distance between all tokens. This prevented our models from learning misleading similarities which would occur when using other embeddings like phoc-encoding.[43]

When comparing SOTA-model performances on mutation "err" (Figure 8) and "syn" (Figure 7), resulting F1-Scores matched for all models except LKE and MoLFI. LKE featured the highest difference when "syn" and "err" mutation results were compared. No general preference regarding adaptability or dependence on the mutation case could be identified for LKE and MoLFI. However, both did not result in competitive F1-Scores throughout different data sets. The differences of LKE and MoLFI to the other SOTA-parsers performances originated from their setup. All SOTA-parsers were retrained for each and every mutation case and starting line combination. However, LKE clustered and detected anomalies differently for both mutation cases. Similarly, MoLFI created different log templates for both mutation cases since it optimized multiple objectives at once. All other SOTA-parsers mainly relied on word frequencies, thus, parsing results only depended on mutation locations (500, 1000, 1500), not mutation case ("syn" and "err").

### 4.2.2 | Parsing flexibility

In a changing world, parsers need to cope with changing log file structures. In the following, we discuss parsing flexibility of the studied SOTA-parsers followed by DL-methods.

Although SOTA-parsers are unsupervised and, thus, event parsing templates were retrained on the mutated files, respectively, they only achieved high parsing performances for individual data sets. Overall, AEL and Drain performed best among SOTA-parsers when applied to log files with and without mutation. Examining SOTA-parser performances in detail, we discovered that AEL and Drain performed despite mutation very well for two data sets, *BGL* and *Linux*. As illustrated in Figure 2B, desired event values from *BGL* start occurring after log line number 1000 only. Thus, parsing templates fit for mutations starting at 500 and 1000 and were identical. This led to F1-Scores of 1.0. Only when mutations were induced starting at line 1500, parsing templates were not flexible enough resulting in drastic decrease of F1-Score down to 0.08 for AEL and Drain, similarly. Furthermore, SOTA parsing performances of data set *Linux* yielded extreme F1-Scores of roughly 0.0 or 1.0. *Linux* logs and the chosen log message of interest exhibited the special property that the event value to be parsed was a string, not a number as in all other data sets. Following, the parsing template was built by Drain and AEL like "session opened for <*> <*> by (uid=<*>)." Thus, the event key was considered as a variable part as well. This made parsing flexible for any event key changes and led to correct parsing results. This demonstrates that parsing performance was independent of the exact event modification for *Linux*. Changing the mutation types would not affect the results since the learnt parsing templates would still fail or entirely generalize.

The studied DL-parsers delivered almost constant high results. In both mutation cases, "syn" (Figure 7) and "err" (Figure 8), the applied DL-parsers performed equally good or only slightly worse compared to parsing results achieved on non-mutated test sets. This supports the underlying hypothesis that DL-parsers are flexible such that they still parse log messages correctly despite gradual changes. All DL-parsers were trained on the entire log message of interest, thus, involved all surrounding tokens for deciding on the desired parsing result. Table 3 provides resulting F1-Scores comparing the non-mutated and mutated cases per data set. In each column the best value for both, "Mutation" equal to yes and no, are marked in bold. All DL-parsers remained constant parsing performance for all data sets undergoing mutation except one data set, *HealthApp*. In that case, parsing of mutated log files reduced F1-Score achieved by stateful LSTM from 0.942 to 0.917 (see Table 3). This can be explained by the nature of log file containing two very similar log messages as shown in Figure 1. One contained "totalCalories," the other held "totalAltitude." As soon as "totalCalories" mutated, the log message could be confused with the "totalAltitude" resulting in "False" indicating that the desired calories were not found to be parsed. This explanation holds for all DL-parsers. Thus, FlexParser works better the more tokens within a log message are static and the more unique a log line is. Parsing the first data set, *Android*, performed best with FCN or GRUFCN, both achieving an F1-Score of 0.995 (see Table 3). All other methods were closely behind with 0.987, where all mutated and non-mutated results shared the same value. We explain this with *Android* exhibiting many variable parts in the chosen log message and relatively low frequency. In this circumstance, convolutional characteristics of the model as used in FCN and GRUFCN contributed to the slight advantage over the other models. For all data sets other than *Android*, the stateful LSTM reached highest F1-Scores in both cases. It even parsed four data sets (*Linux*, *Mac*, *Spark*, *Windows*) perfectly. The lowest F1-Score of 0.719 was found for *Windows* using LSTMFCN. Furthermore, LSTMFCN as well as LSTM were outperformed by other DL-methods for any tested data set. The stateful LSTM achieved highest median F1-Score throughout all our flexible parsing experiments (see Figures 7 and 8). We attribute the advantage of stateful LSTM to the properties of statefulness. This ensured that patterns across several samples were learnt which addresses the needs of log file parsing, perfectly.

Concluding, results for both mutation cases were very similar. Figures 7 and 8 illustrate that SOTA-parsers fell behind DL-parsers drastically in case of mutation. This resulted in all DL-parsers achieving higher mean and median F1-Scores compared to any SOTA-parser. Among all models, FlexParser representing the stateful LSTM proved flexibility and resulted in best F1-Scores.

**TABLE 3** F1-Scores of all DL-parsers per data set

| Method | Mutation | Android | BGL | HealthApp | Linux | Mac | Spark | Windows |
|---|---|---|---|---|---|---|---|---|
| FCN | no | **0.995** | 0.984 | **0.942** | **1.000** | 0.999 | 0.920 | **1.000** |
| FCN | yes | **0.995** | 0.984 | 0.896 | **1.000** | 0.999 | 0.920 | **1.000** |
| GRUFCN | no | **0.995** | 0.984 | **0.942** | 0.998 | **1.000** | 0.922 | **1.000** |
| GRUFCN | yes | **0.995** | 0.984 | 0.896 | 0.998 | **1.000** | 0.922 | **1.000** |
| LSTM | no | 0.987 | 0.984 | 0.936 | 0.939 | 0.993 | 0.916 | 0.922 |
| LSTM | yes | 0.987 | 0.984 | 0.896 | 0.939 | 0.993 | 0.916 | 0.922 |
| stateful LSTM | no | 0.987 | **0.985** | **0.942** | **1.000** | **1.000** | **1.000** | **1.000** |
| stateful LSTM | yes | 0.987 | **0.985** | 0.917 | **1.000** | **1.000** | **1.000** | **1.000** |
| LSTMFCN | no | 0.987 | 0.984 | 0.888 | 0.982 | 0.993 | 0.999 | 0.719 |
| LSTMFCN | yes | 0.987 | 0.984 | 0.888 | 0.982 | 0.993 | 0.999 | 0.719 |

*Note:* Results of parsing test sets which contain the same event keys and structure as the training set (Mutation = no) are compared with results of parsing test sets which exhibit mutated event keys (Mutation = yes). Bold entries highlight the best results per data set for both, mutated and non-mutated case.

## 4.3 | Best model comparison

In all our experiments, Drain represented the best SOTA-parser whereas stateful LSTM performed best among studied DL-parsers. In the following, we provide deeper insights into performance differences per data set for those two methods. Figure 9 holds F1-Scores per data set for the non-mutated next to the "syn" case for comparison. All bar charts illustrate the F1-Score for Drain in light gray and for stateful LSTM in dark gray side by side per data set. Since performance of those two methods were identical when comparing "syn" and "err," only details for "syn" case are presented.

First, we discuss the non-mutated case. Figure 9A displays perfect parsing capabilities for both methods when applied to non-mutated test sets of *Linux*, *Mac*, *Spark* and *Windows*. For *BGL* and *HealthApp*, Drain still performed perfectly exhibiting an F-Score of 1.0. Stateful LSTM fell behind slightly with 0.98 and 0.94, respectively. Here, the same explanation applies as before. These data sets contained very similar log message types which got confused in rare occasions by stateful LSTM, whereas Drain could apply the exact event template reliably. On the contrary, stateful LSTM parsed almost all values of *Android* correctly (F1-Score of 0.99), whereas Drain experienced major difficulties and resulted in only 0.14 F1-Score. We explain the low performance of Drain when applied to *Android* by the log message properties. The chosen to be parsed event key was logged in *Android* together with many other variables in the same log message. With the default configuration of Drain, templates were extracted which depend on the simultaneous occurrence of other variables holding specific values which was only true for a small fraction of log lines. Thus, for better performance of Drain, thresholds needed to be adapted manually which interfered with the need for automation.

When inducing mutations and applying the chosen DL-parser, the high performance remained identical in almost all cases. Only *HealthApp* showed slight deviation of two percent at most down to 0.92 in Figure 9B. Here again, mutation amplified the confusion of two similar log lines. The decisive event key "totalCalories" was contained in a log message which was very similar to the very next line containing the same structure but logging "totalAltitude" (see Figure 1). For all other datasets, the DL-parser delivered the same good results. Drain performed worse and differently depending on both, the used data set and line number the mutation starts. When testing Drain on the mutated case starting at line 500, only *BGL* and *Linux* could be parsed correctly. All other data sets resulted in only 55% F1-Score and below. The performance increased again if the test data set contained equal amounts of mutated and non-mutated rows (see Figure 9C). As soon as mutated lines shared larger proportion (Figure 9D), Drain performed better again with some exceptions. For example, performance of parsing *BGL* dropped drastically down to 0.08, while *Android* remained low. The low performance originated from the dependence of Drain on the number of present, equally constructed log messages. As soon as the mutated rows represented the majority of log lines in training, the parsing template was updated to the mutation case and could parse the mutated cases correctly but failed for the first, non-mutated lines. Furthermore, performance differences across data sets could be explained by the different distribution of decisive log messages as shown in Figure 2.

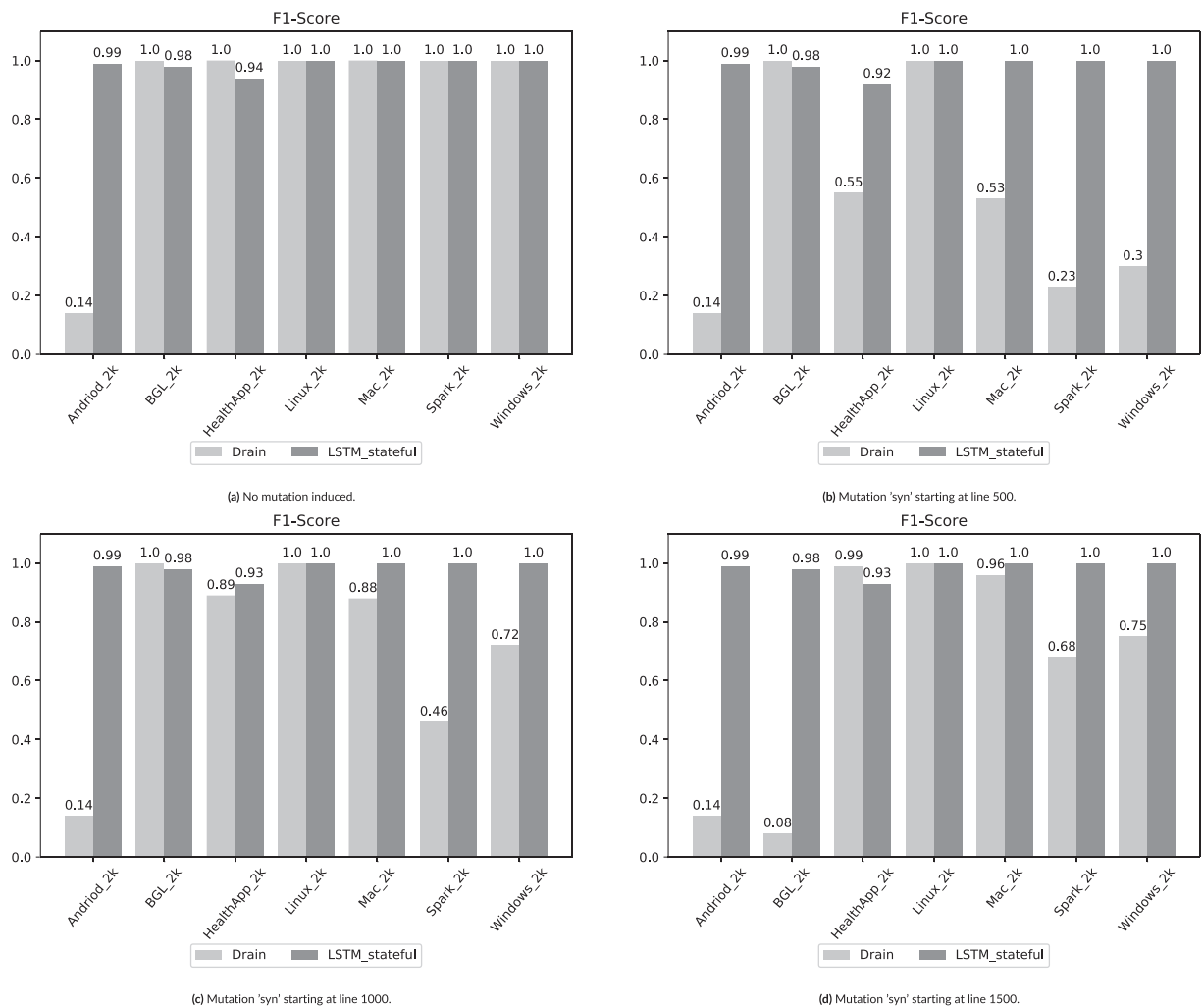

**FIGURE 9** Parsing performance per data set for mutation case "syn" induced starting in different line numbers. F1-Scores are displayed for the best performing SOTA-parser (Drain) and DL-parser (stateful LSTM) next to each other

Concluding, the stateful LSTM outperformed Drain as soon as log files change and undergo mutation. Depending on the tested data set and its distribution of decisive log messages, Drain parsed reliably (e.g., for *BGL*) or decreased drastically (e.g., for *Spark*). Thus, for applications where a parser needs to perform well for different data sets, stateful LSTM poses the method to choose.

Finally, we propose a new, flexible parser called *FlexParser*. It comprises a DL pipeline which contains preprocessing of raw log file text, vectorization using one-hot-encoding and supervised learning of a stateful LSTM. This model is trained on the desired event parsing logic per event key. We demonstrated that it parses desired values correctly for various data sets, because in stateful models the parsing rules are learnt across several examples. Furthermore, flexibility is pointed out by applying the method onto mutated log file lines which still results in correctly parsed values for predominant cases. This flexibility originates from the FlexParser using the entire message as input and not relying on single parsing triggers which might be subject to mutation. Another advantage is that the method needs to be trained on a comparably small amount of log messages. If the method is regularly re-trained, then also continuous log changes do not affect parsing performance.

## 5 | DISCUSSION

We studied parsing performance of several models with focus on their flexibility in case of log message changes. In the following, we discuss potential threats to the validity of our study followed by limitations of our approach.

Our study contains real world data combined with synthetically induced mutations. Potential threats to the validity of our parsing performance study are representativeness of the log data and model comparison methodology. We included in our study different mitigations in order to guarantee general applicability and validity of our results. To address the former, we chose a range of real world log data of various kinds of systems as presented in Table 1. Moreover, we selected decisive events per data set which are common for the respective domain. In addition, the event frequency (see Table 2) and distribution (see Figure 2) also reflect great diversity. This leaves us with a diverse set of log file types created by a range of logging mechanisms which we study our approaches on. We believe that they represent software log files of current systems well. All individual parsing performance results contribute equally to the overall average performance. Furthermore, we discuss our model comparison methodology. Our goal was to find and demonstrate the best suitable model for flexible parsing. We mitigated threats to validity by choosing F1-Score as the decisive performance metric. F1-Score was chosen over AUC (Area under the receiver operating characteristic Curve) and accuracy out of the following reasons. For model comparison, Lobo et al[44] demonstrated that AUC is misleading for model comparison. Furthermore, AUC cannot be calculated for SOTA-parsers. In contrast, F1-Score holds a meaningful performance metric for both, SOTA- and DL-parsers, and poses a popular performance metric in NLP.[45] Furthermore, F1-Score is the harmonic mean of precision and recall, thus, to be chosen over accuracy for imbalanced data like ours. Another potential threat to the validity of our model comparison originates from SOTA-parsers being unsupervised and DL-parsers supervised methods. In order to ensure comparability of the best possible parsing performances per method, we applied SOTA-parsers from scratch and retrained them for all mutation cases for all data sets each. This led to possibly adapted regular expressions for all studied mutation cases and leveraged the method's full potential.

Although our proposed method, FlexParser, exhibits great flexibility, we discuss some limitations in the following. First, we did not study insertion nor deletion of logging code. This is because our focus was to create stable, high-quality parsing results for reliable subsequent analysis. Thus, additional variables being inserted to logging messages will not deteriorate the stability. In contrast, deletion of variables from logging can have tremendous effects on further analyses. Nevertheless, we did not cover those in our studies since deletions of log messages are typically found in less than 2% of all logging modifications.[7] Another limitation is found in our experimental design. Currently, our model is set up to be trained separately for any desired event value to be parsed. This ensured comparability across data sets, models, mutation cases, and mutation insertion locations. Now that we found the best model for flexible parsing, in future research parsing capabilities for several, different types of events in the same run shall be investigated. For that, training and respective dimensions need to be adapted. Since we discovered in this study the stateful LSTM as the best model to flexibly parse log files, in successive evaluations the parsing capabilities shall be scaled to several key-value pairs. Furthermore, we only examined gradual changes affecting only one string per log message. This assumption does not always hold in practice. Consequently, the next step should contain studying the turning point of how many changes at a time in the same log message could be digested and which countermeasures can be applied. In future studies, we further propose to combine *FlexParser* with the power of SOTA-parsers in order to leverage benefits of both. Since our presented approach is built for gradually changing log files, a promising SOTA-parser, e.g. Drain, could be utilized for sanity checks. Moreover, the proposed *FlexParser* requires available ground truth for the first training as the stateful LSTM poses a supervised Deep Learning method. To overcome this condition as well, again SOTA-parsers should be utilized to create necessary inputs for further reduction of manual processing. Lastly, anomaly detection techniques should be considered to determine meaningful points in time for retraining our model.

# 6 | CONCLUSION

In this paper, we present a flexible log message parser called *FlexParser*. Our approach extracts reliably event values from gradually changing log messages. We obtain F1-Scores between 0.92 and 1.0 depending on the tested data set. It outperforms all considered state-of-the-art parsing methods as soon as log messages undergo mutations. *FlexParser* consists of text preprocessing steps followed by a stateful LSTM. It reaches higher parsing performances than other four DL-parsers as statefulness forces the model to learn parsing event logic rather than direct classification relations. For comparability and demonstration purposes, we chose one event per data set to be parsed. For general applicability, we tested on seven different log file data sets written by distributed systems, supercomputer, operating systems and mobile systems. We compared results achieved by DL-parsers with those from eleven different SOTA-parsers. We specifically studied parsing performance on log files which are subject to gradual changes. Two different kinds of those changes, called mutations, have been induced. Furthermore, we parsed log files which contain mutations starting at three different line numbers. Our results show that *FlexParser* adapts to gradual changes and parses desired values correctly because it takes contents of the entire log message into account, not only a single decisive trigger. To our knowledge, the proposed method is the first adaptive parser which does not rely on continuity in semantics. Furthermore, our approach does not only detect mutations but performs correct parsing despite existing mutations. Thus, *FlexParser* poses the first method to holistically and successfully parse values from gradually changing log messages. By proving successful parsing for a variety of data sets, we could address this research gap. In future research, our model should be expanded to several parsing events at once. Furthermore, *FlexParser* shall be combined with other existing parsers to extend automation and robustness.

### ACKNOWLEDGEMENT
We are grateful to Log Analytics Powered by AI (logPAI) for providing access to several log file data bases and code of traditional, pattern-based log file parsers. Open Access funding enabled and organized by Projekt DEAL.

### DATA AVAILABILITY STATEMENT
The data that support the findings of this study are openly available in zenodo at https://doi.org/10.5281/zenodo.3227177, reference number 3227177.

### ORCID
*Nadine Rücker*  https://orcid.org/0000-0001-8157-1689